\pdfoutput=1

\documentclass[conference]{IEEEtran}
\ifCLASSINFOpdf
\else
\fi
\ifCLASSOPTIONcompsoc
  \usepackage[caption=false,font=normalsize,labelfont=sf,textfont=sf]{subfig}
\else
  \usepackage[caption=false,font=footnotesize]{subfig}
\fi
\usepackage{graphicx}
\usepackage{amsmath}
\usepackage{amssymb}
\usepackage{algorithmic}
\usepackage{algorithm}
\usepackage{latexsym}
\usepackage{array}
\usepackage{mdwmath}
\usepackage{mdwtab}
\usepackage{eqparbox}
\usepackage{bm}
\usepackage{mathtools}
\usepackage{multirow}
\usepackage{booktabs}
\usepackage{xcolor}
\usepackage{cases}
\usepackage[leftcaption]{sidecap}
\usepackage{harpoon}
\usepackage{wrapfig}
\usepackage{cite}
\usepackage{epstopdf}
\usepackage{amssymb}

\usepackage{graphicx}
\DeclareMathOperator*{\argmax}{arg\,max}
\DeclareMathOperator*{\argmin}{arg\,min}
\begin{document}
%
\title{Detecting Drivable Area for Self-driving Cars: \\An Unsupervised Approach }

\author{
\IEEEauthorblockN{
Ziyi Liu,
Siyu Yu,
Xiao Wang
and Nanning Zheng\IEEEauthorrefmark{1}}
\IEEEauthorblockA{Institute of Artificial Intelligence and Robotics, Xi'an Jiaotong University, Xi'an, Shannxi, P.R.China\\
National Engineering Laboratory for Visual Information Processing and Applications,\\
Xi'an Jiaotong University,
Xi'an, Shannxi, P.R.China\\
Correspondence : \IEEEauthorrefmark{1}nnzheng@mail.xjtu.edu.cn
}}


\maketitle

\begin{abstract}
It has been well recognized that detecting drivable area is central to self-driving cars. Most of existing methods attempt to locate road surface by using lane line, thereby restricting to drivable area on which have a clear lane mark. This paper proposes an unsupervised approach for detecting drivable area utilizing both image data from a monocular camera and point cloud data from a 3D-LIDAR scanner.
Our approach locates initial drivable areas based on a "direction ray map" obtained by image-LIDAR data fusion.
Besides, a fusion of the feature level is also applied for more robust performance.
Once the initial drivable areas are described by different features, the feature fusion problem is formulated as a Markov network and a belief propagation algorithm is developed to perform the model inference.
Our approach is unsupervised and avoids common hypothesis, yet gets state-of-the-art results on ROAD-KITTI benchmark. Experiments show that our unsupervised approach is efficient and robust for detecting drivable area for self-driving cars.

\end{abstract}


%
\IEEEpeerreviewmaketitle

\section{Introduction}
\begin{figure}[t]
  \centering
  \includegraphics[width=0.48\textwidth]{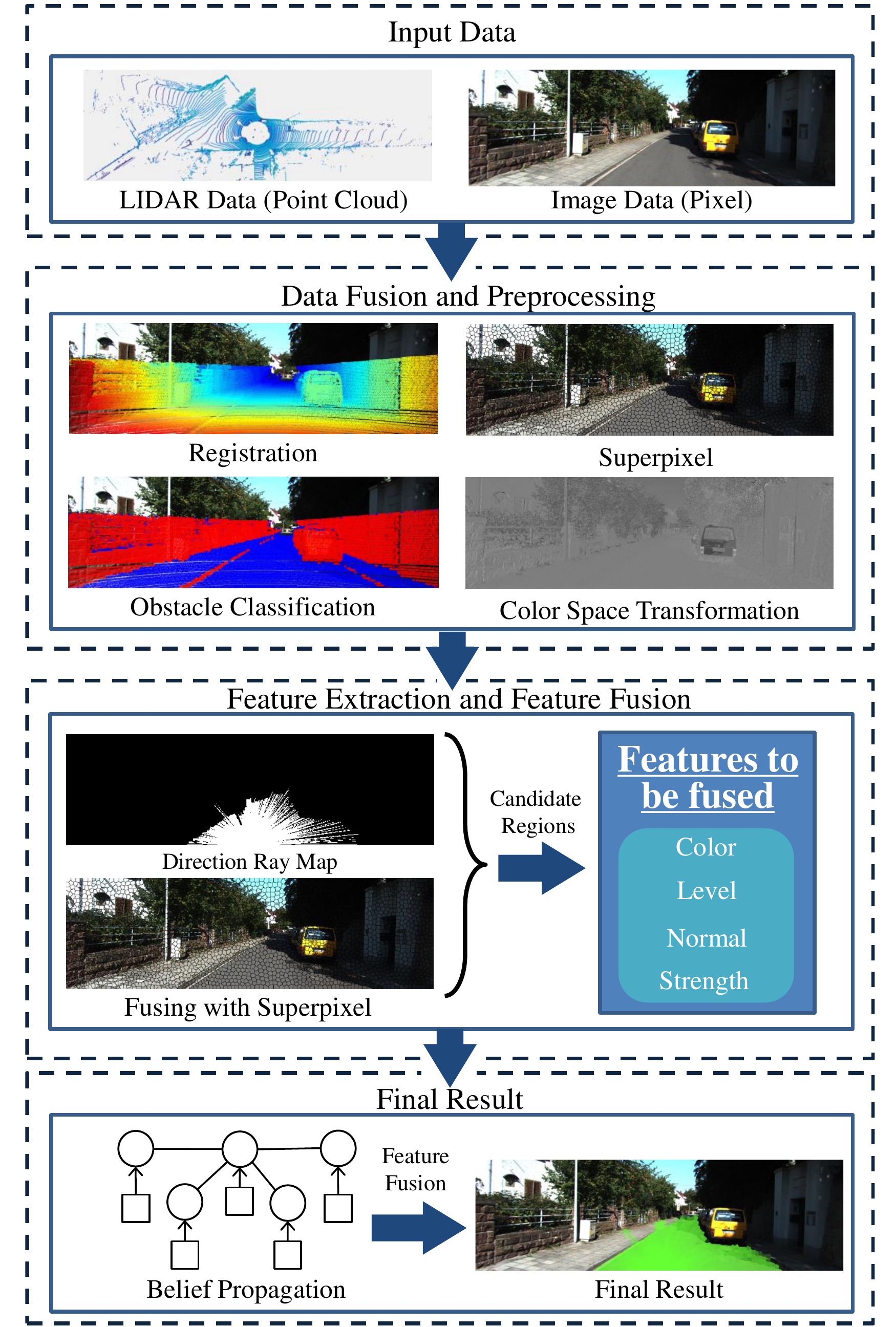}\\
  \caption{The framework of our proposed unsupervised approach.}\label{fig:framework}
\end{figure}
In the field of self-driving cars, road detection is a crucial requirement, and the topic has been attracting considerable research interest in recent times. Moreover, significant achievements on road detection have been proposed in the literature\cite{bar2014recent}. Though there exist some algorithms on road detection for well-marked roads based on sample-training, unsupervised road detection for unlabeled roads in inner-city and rural areas still remains a challenge on account of the high variability of traffic scene and light conditions. So far, to the best of our knowledge, there exists no robust solution to solve this challenge. However, taking cues from human driving behavior, distinguishing drivable area from non-drivable area is a priority for humans when they drive. Then, roads are found and driving decisions are made based on the drivable area.

Inspired by human driving behavior, we proposed an unsupervised approach for detecting drivable area by fusing image data and LIDAR points, as shown in \figurename\ref{fig:framework}.
By combining image coordinate frame with LIDAR coordinate frame, a Delaunay triangulation \cite{lee1980two} is generated to describe the spatial-relationship between points and utilized to classify obstacle points.
Then an initial location of the drivable area is obtained by the fusion of "direction ray map" and image superpixels, which serves as priori knowledge and narrows the range of detection, as detailed in Section \uppercase\expandafter{\romannumeral3}.
In Section \uppercase\expandafter{\romannumeral4}, features used to describe the final drivable area are learned autonomously based on that initial location.
In Section \uppercase\expandafter{\romannumeral5}, a feature fusion step is implemented leveraging a Markov network through belief propagation, and the final results are obtained.

In the experiment step, our approach is tested on ROAD-KITTI benchmark\cite{Fritsch2013ITSC}. Comparisons have been made with similar fusion approaches and ours gets state-of-the-art results without training or making assumptions about shape or height, which demonstrates the robustness and generalization ability of our approach.

\section{Related Work}
Reliably detecting the road areas is a key requirement in self-driving cars. In resent years, many approaches have been proposed to address this challenge.
The approaches mainly differ from each other based on the type of sensors used to get data, such as, monocular camera\cite{Kuehnl2012ITSC}, binocular camera\cite{badino2007free} , LIDAR\cite{tongtong20113d} and the fusion of multi-sensor\cite{xiao2015crf}.

For monocular camera based approaches, most road detection algorithms use cues such as color \cite{broggi1995vision} and lane markings\cite{nan2016efficient}.
To cope with illumination varieties and shadows, different color spaces have been introduced \cite{jau2008comparison}\cite{alvarez2011road}\cite{maddern2014illumination}.
Besides, leveraging deep learning, monocular vision based methods can achieve unprecedented results\cite{badrinarayanan2015segnet2}\cite{alvarez2012road}\cite{OB16b}.
However, unlike other vision conceptions, such as cat or dog, the conception of a road cannot be defined by appearance alone. A region that is regarded as a road depends more on its physical attributes.
Therefore, approaches only relying on monocular vision are not robust enough for real applications.

With the advent of LIDAR sensors, which can measure distances accurately, many LIDAR-based road detection approaches have been developed.
Such approaches use the LIDAR points' spatial information to analyse the scene and regard flat areas as roads.
But due to the sparsity of LIDAR points, it's hard to analyse the details between points.
Besides, abandoning image information will increase the difficulty of points classification.

Considering the above drawbacks of the above methods, we propose an unsupervised detection approach for drivable area by fusing image data and LIDAR points.
Compared with other detection methods, the superiorities of our approach reflect in three aspects.
First, it dose not need strong hypothesis, training steps or manually labelled data, which ensures the generalization ability of our approach.
Second, by fusing LIDAR and monocular camera, our approach can learn probabilistic models in a self-learning manner, thereby making it robust to complex road scenarios and fickle illumination.
Finally, superpixels are used as basic processing units instead of pixels, and they have been found to be an easy and efficient way to combine sparse LIDAR points with image data.
Therefore, we consider our unsupervised approach as an efficient and robust way for drivable areas detection for self-driving cars.

\section{Preprocessing and Data Fusion}

\subsection{Image Processing in Superpixel Scale}
In our approach, superpixels are considered as elementary units instead of pixels motivated by the observation that superpixels and LIDAR points are complementary.
On the one hand, superpixels are dense and involve color information that LIDAR sensor cannot capture; on the other hand, LIDAR points reflect depth information that monocular camera cannot obtain.
Owing to the promotion of superpixel segmentation methods, image processing in superpixel scale can cut down the computational and memory consumptions without much loss in accuracy.
In our approach, "Sticky Edge Adhesive Superpixels" are detected and used\cite{dollar2013structured}\cite{DollarICCV13edges}\cite{ZitnickECCV14edgeBoxes}.
Without edge term, these superpixels are computed using an iterative approach like SLIC\cite{achanta2012slic}; with edge term added, the superpixels snap to edges, resulting in higher quality boundaries.
So it can be assumed that superpixels can adhere object boundary well.
Therefore, it's a reasonable choice to use superpixels to shape the initial location of drivable areas, which will be detailed in \uppercase\expandafter{\romannumeral4}.

Besides, superpixels can be used to calculate local statistics so that results can be more robust.
Thus, in our approach, superpixels are considered as elementary units instead of pixels, and assist in shaping the initial drivable area and accelerating feature extraction processes.

\subsection{Illumination Invariant Color Space}
To acquire color feature regardless of lighting condition or weather, RGB color images ($I$) are transformed into an illumination invariant color space, noted as $I_{ii}$. As presented in \cite{maddern2014illumination}, a 3-channel image is converted into a 1-channel image with one parameter $\alpha$ related to peak spectral responses of the camera:
\begin{equation}\label{log}
\frac{1}{{\lambda}_2}=\frac{\alpha}{{\lambda}_1}+\frac{(1-\alpha)}{{\lambda}_3}
\end{equation}
where ${\lambda}_1$,${\lambda}_2$,${\lambda}_3$ are wavelengths and the $I_{ii}$ is obtained following:
\begin{equation}\label{log}
\begin{split}
{I}_{ii}(u,v)=log({I_G(u,v)})-{\alpha}log({I_R(u,v)})\\
-(1-\alpha)log({I_B(u,v)})
\end{split}
\end{equation}
where ${I}_{ii}(u,v)$ is the pixel value of ${I}_{ii}$ in $(u,v)$ and $I_R(u,v)$, $I_G(u,v)$, $I_B(u,v)$ are $R$, $G$, $B$ pixel values of ${I}$ in $(u,v)$, respectively. We set $\alpha$ to 0.4706 as\cite{maddern2014illumination} suggested.

\subsection{Obstacle Classification via Data Fusion}
\begin{figure}[t]
\centering
  \includegraphics[width=0.49\textwidth]{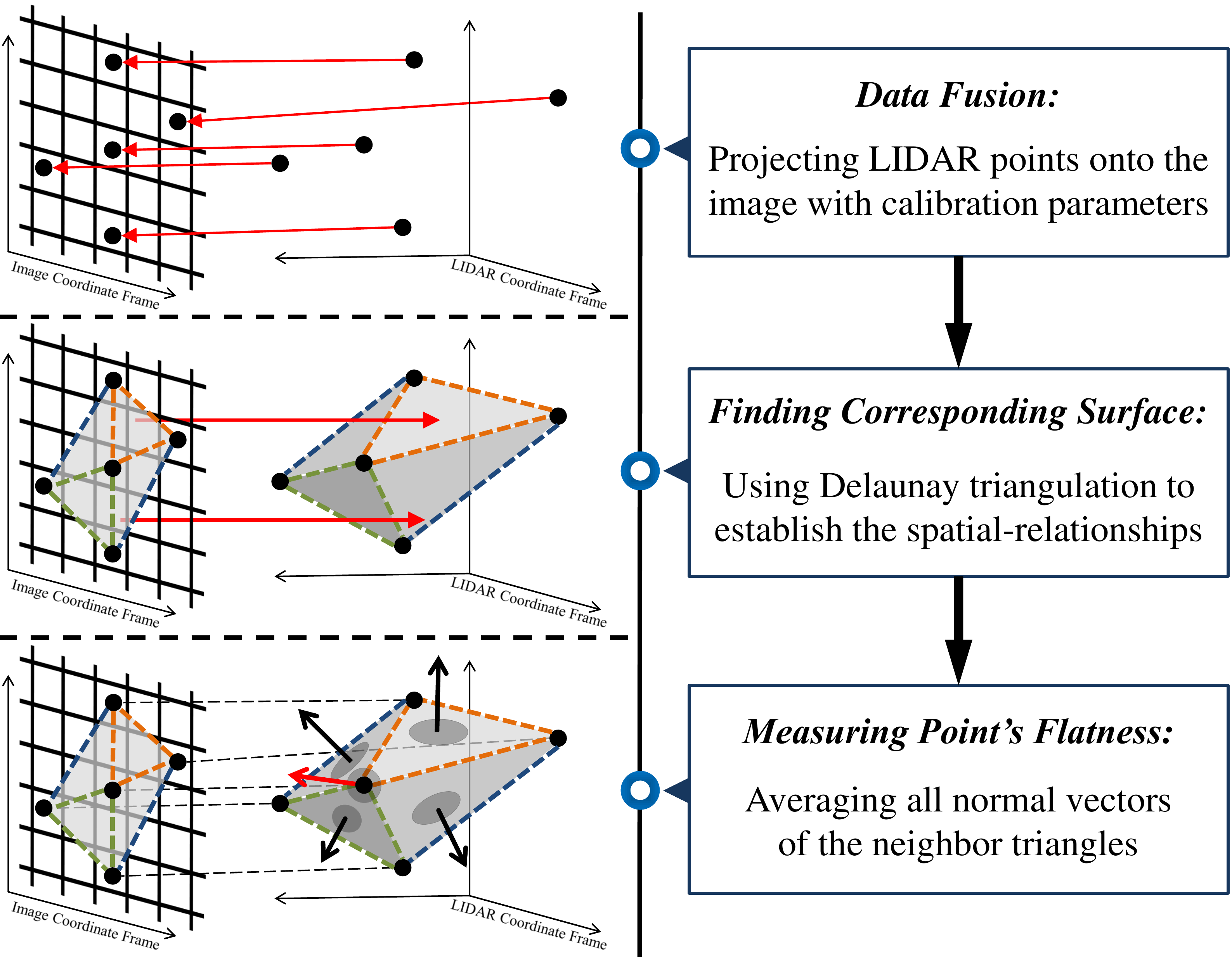}\\
  \caption{The whole process of obstacle classification, which contains three steps: data fusion, finding corresponding surface and measuring point’s flatness. The red arrows indicate the main information obtained by each step.}\label{fig:normal}
\end{figure}
The whole process of this subsection is shown as \figurename\ref{fig:normal}. To fuse LIDAR points with image pixels, the projection of 3D points is employed as presented in \cite{Geiger2013IJRR}. After the alignment, the LIDAR points set noted as $\mathbb{P}=\{\mathbf{P}_i\}_{i=1}^N$ is gained, where $\mathbf{P}_i=(x_i,y_i,z_i,u_i,v_i)$. $(x_i,y_i,z_i)$ is $\mathbf{P}_i$'s LIDAR coordinate and $(u_i,v_i)$ is its image coordinate.

\begin{figure}[t]
\centering
\subfloat[]{\includegraphics[width=0.47\textwidth]{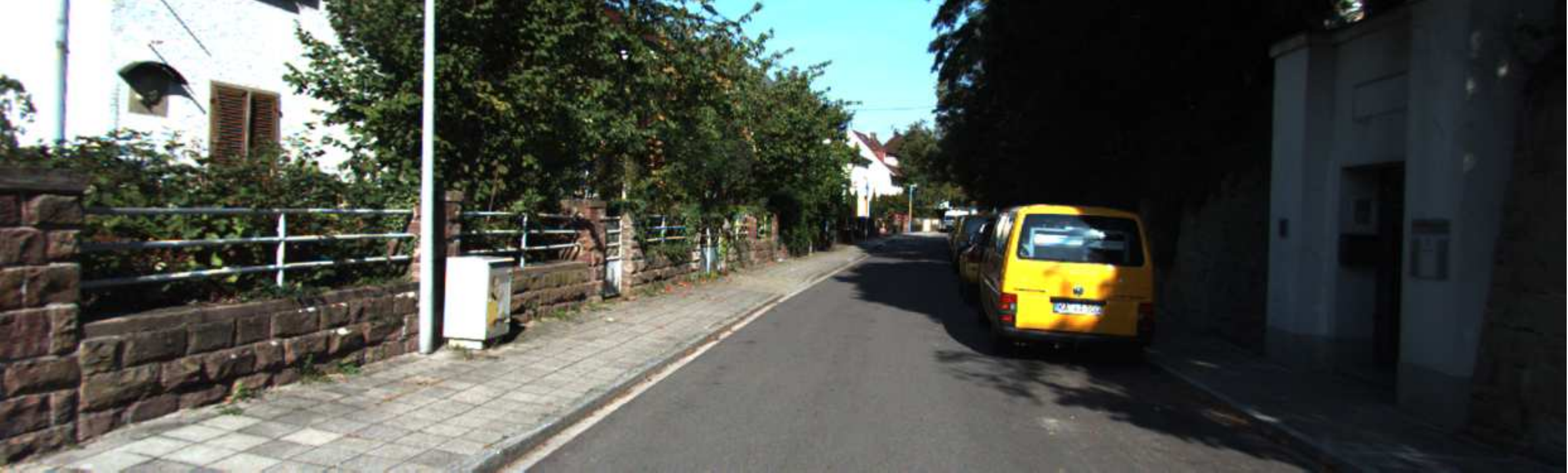}}
\hfil
\subfloat[]{\includegraphics[width=0.47\textwidth]{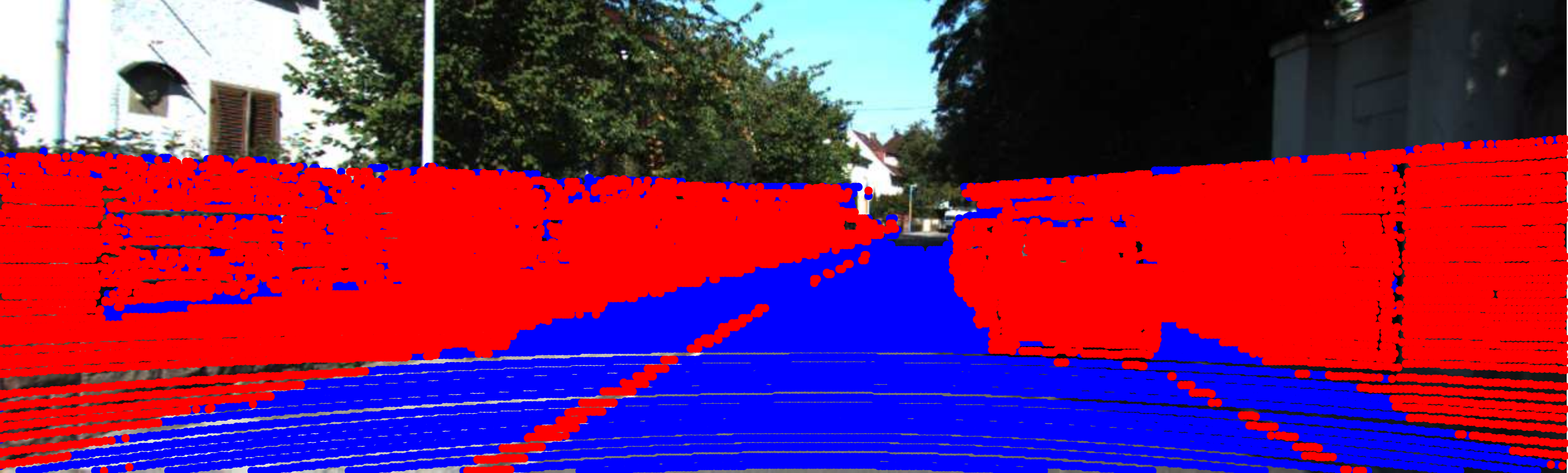}}
\hfil
\caption{The classification result of obstacle and non-obstacle. (a) is the original image. (b) shows obstacle points (red dots) and
non-obstacle points (blue dots). It can be seen that LIDAR points are successfully classified.}
\label{fig:obstacle}
\end{figure}

The objective of obstacle classification step is to find the mapping relations $ob(\mathbf{P}_i)$.
$ob(\mathbf{P}_i)=1$ indicates that $\mathbf{P}_i$ is an obstacle point while $ob(\mathbf{P}_i)=0$ indicates that $\mathbf{P}_i$ is a non-obstacle point (as shown in \figurename\ref{fig:obstacle}).
It is assumed that whether $\mathbf{P}_i$ is an obstacle point or not merely depends on how flat its corresponding surface is in the physical world, so the problem is broken up into two sub-problems: how to define the corresponding surface of $\mathbf{P}_i$ and how to measure the its flatness, as shown in \figurename\ref{fig:normal}.

To define the corresponding surface of $\mathbf{P}_i$, Delaunay triangulation is utilized to establish the spatial-relationships among $\mathbb{P}$, because it has properties that each vertex has on average six surrounding triangles in the plane, and nearest neighbor graph is a subgraph of the Delaunay triangulation.
The graph is generated as proposed in \cite{6856454}.
For each $\mathbf{P}_i$, its image coordinate $(u_i,v_i)$ is used in a planar Delaunay triangulation to generate an undirected graph $G = \{ \mathbb{P},E\} $.
$E$ represents the set of edges which defines the relationships among $\mathbb{P}$.
The edge $(\mathbf{P}_i,\mathbf{P}_j)$ is discarded if it dose not satisfy :
\begin{equation}
\left\| {\mathbf{P}_i - \mathbf{P}_j} \right\| < \varepsilon
\end{equation}
where $\left\| {\mathbf{P}_i - \mathbf{P}_j} \right\|$ is the Euclidean distance of $({x_i},{y_i},{z_i})$ and $({x_j},{y_j},{z_j})$.

Then, the "corresponding surfaces" are the surfaces (triangles) determined by $\{(u_j,v_j)~| ~j=i~or~\mathbf{P}_j \in Nb(\mathbf{P}_i) \}$, where $Nb(\mathbf{P}_i)$ is the set of $\mathbf{P}_i$'s neighbor points.
Then, the flatness of $\mathbf{P}_i$'s corresponding surfaces can be measured by calculating the normal vectors of them, and $N({{\bf{P}}_i})=(x_i^n,y_i^n,z_i^n)$ is obtained by averaging the normal vectors of $\mathbf{P}_i$'s neighboring triangles. Finally, $ob(\mathbf{P}_i)$ is gained:
\begin{equation}
ob(\mathbf{P}_i) =
  \begin{cases}
    1       & \quad \text{if } arcsin(\frac{z_i^n}{\left\|{N({{\bf{P}}_i})}) \right\|})> c\\
    0       & ~~~~~~~~~~~~~~~~~~~otherwise\\
\end{cases}\
\end{equation}
where $c$ is the minimum deviation angle from horizon of an obstacle point.

\section{Drivable Areas Detection}

To locate the drivable area, we first get an initial location of it and then truing it by features, which is a coarse to fine process.
\subsection{Locating Initial Drivable Areas}

\begin{figure*}[]
\centering
\subfloat[]{\includegraphics[width=0.325\textwidth]{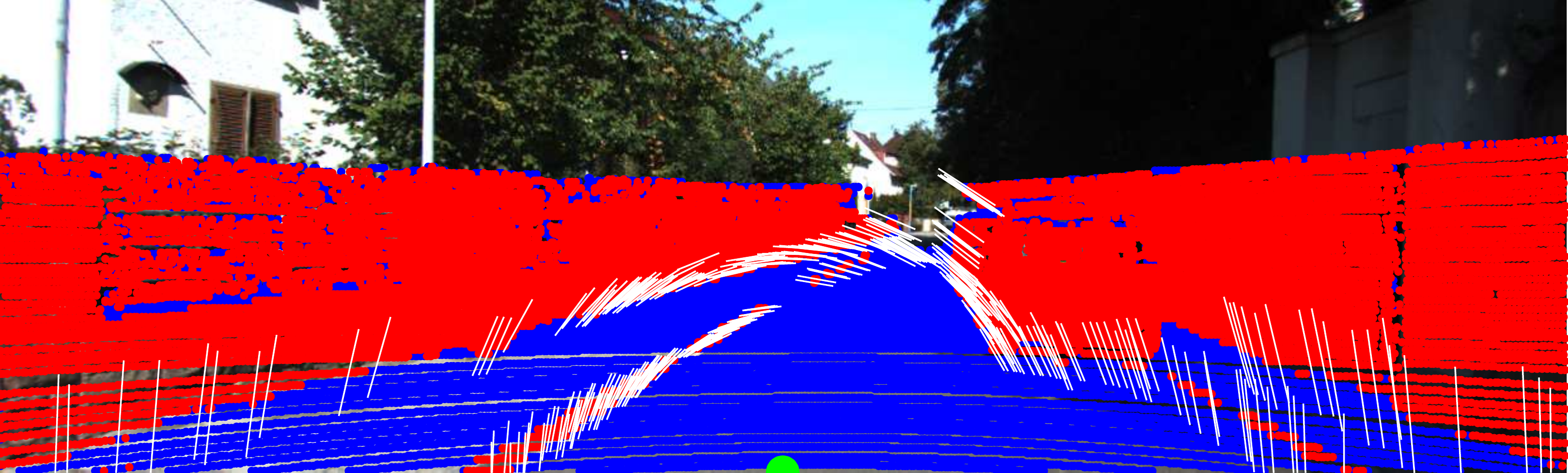}
\label{fig_first_case}}
\hfil
\subfloat[]{\includegraphics[width=0.325\textwidth]{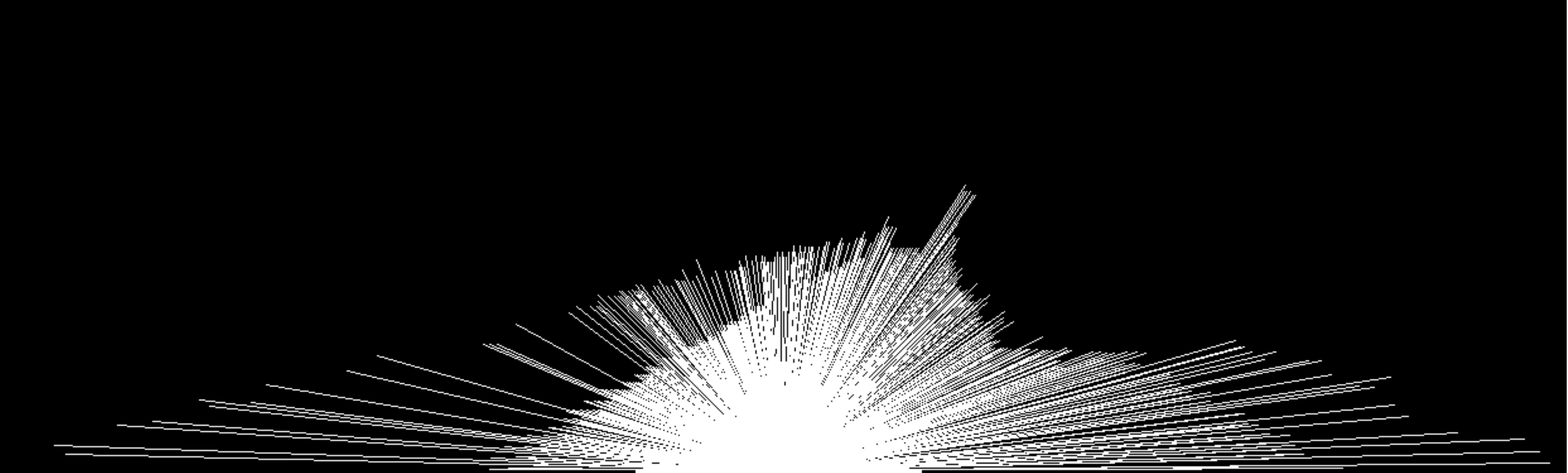}
\label{fig_second_case}}
\hfil
\subfloat[]{\includegraphics[width=0.325\textwidth]{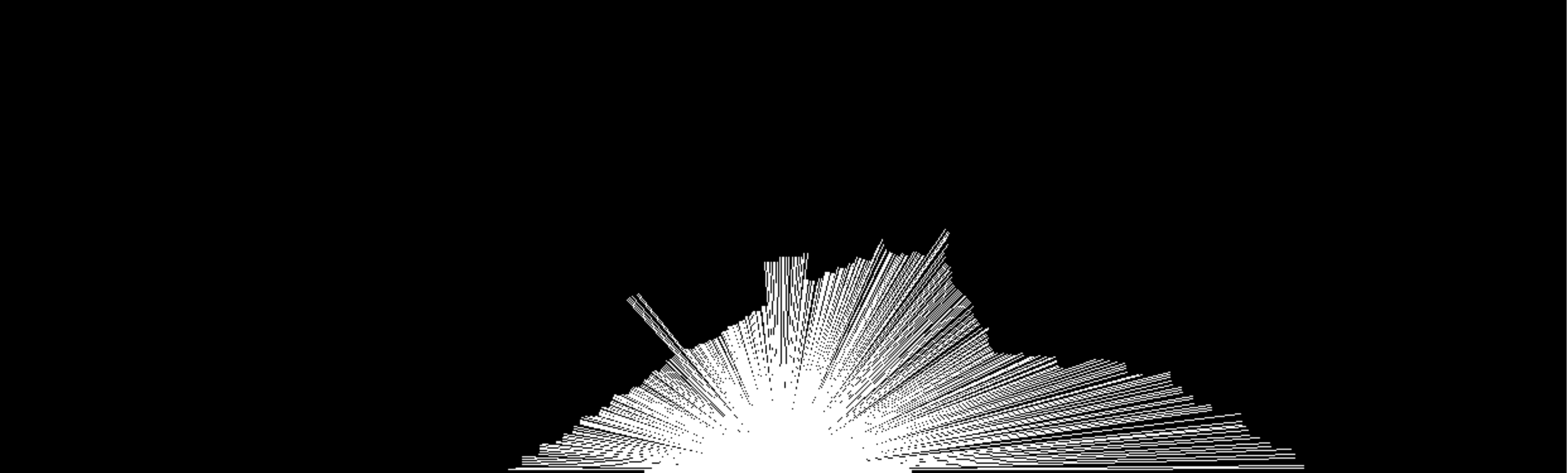}
\label{fig_first_case}}
\hfil
\caption{"Ray leak" problem. (a) shows the "ray leak" problem above the obstacle classification result. Every white line in (a) is perpendicular to a ray and the middle point of white line is the end point of a ray. (b) shows all rays before filtered with white lines. (c) is the result of minimum length filtering}\label{fig:leak}
\end{figure*}

Once the classification step is completed, the corresponding drivable areas are determined.
To locate initial drivable areas, we first get the "direction ray map ($I_{DRM}$)", and then fuse it with superpixels.
$I_{DRM}$ is obtained as shown in Algorithm.\ref{alg:DRM}.
First, polar coordinate transformation is employed to $(u_i,v_i)$, taking the middle bottom pixel of $I$ as the origin, noted as $\mathbf{P}_{base}$.
Then, $\mathbb{P}$ is restructured as $\{\mathbb{P}^{(h)}\}_{h=1}^{H}$ where $\mathbb{P}^{(h)}=\{\mathbf{P}_i^{(h)}\}_{i=1}^{N^{(h)}}$.
$\mathbf{P}_i^{(h)}$ means a LIDAR point whose transformed image coordinate is in the $h$-th angle range.
Because $\mathbb{P}$ is sparse in laser coordinate frame, two problems emerge: the first is how to transform the sparse rays into dense pixels; the second is how to overcome the "ray leak" problem shown in \figurename\ref{fig:leak}.

As for the first problem, increasing $H$ is a natural solution, meanwhile, it aggravates the "ray leak" problem. Therefore, $I_{DRM}$ is fused with superpixels to address this problem.

For the second problem, it should be noted that the width of a vehicle is not ignorable. That is, whether an area is drivable or not depends on the flatness of the area, as well as its width. Therefore, a minimum length filtering method is used to filter the leaked rays, and the final $I_{DRM}$ is shown in \figurename\ref{fig:leak}(c).

Once $I_{DRM}$ has been obtained, the initial drivable area is generated by fusing $I_{DRM}$ with superpixels. Essentially, it is a set of superpixels noted as $\mathbb{S}_{int}=\{{S}_i~|~{S}_i \bigcup I_{DRM} \neq \varnothing\}$, and the set of LIDAR points within $S_i$ is noted as $\mathbb{P}_{S_i}$.

\begin{algorithm}[b]
\caption{ Generating Direction Ray Map.}
\label{alg:DRM}
\begin{algorithmic}[1] 
\REQUIRE ~~\\ 
The set of LIDAR points $\{\mathbb{P}^{(h)}\}_{h=1}^{H}$;\\
\ENSURE ~~\\ 
Direction Ray Map, $I_{DRM}$ 
\STATE Initial $I_{DRM}$ with the size of $I$ and zeros elements
\FOR{$h=1$ to $H$}
\STATE Find obstacle set $\mathbb{O}=\{\mathbf{P}_i ~|~ob( \mathbf{P}_i)=1 ~\& ~\mathbf{P}_i  \in \mathbb{P}^{(h)} \}$
\IF{$\mathbb{O}=\varnothing$}
\STATE $\mathbf{P}_{bin}^{(h)}=\argmax_{ \mathbb{P}^{(h)} }{dist(\mathbf{P}_i^{(h)},\mathbf{P}_{base})}$
\ELSE
\STATE $\mathbf{P}_{bin}^{(h)}=\argmin_{ \mathbf{P}_i^{(h) } \in \mathbb{O} }{dist(\mathbf{P}_i^{(h)},\mathbf{P}_{base})}$
\ENDIF
\STATE Line point $\mathbf{P}_{bin}^{(h)}$ with point $\mathbf{P}_{base}$ in $I_{DRM}$
\ENDFOR
\end{algorithmic}
\end{algorithm}

\begin{algorithm}[b]
\caption{ Getting "Level" Feature.}
\label{alg:level}
\begin{algorithmic}[1] 
\REQUIRE ~~\\ 
The set of LIDAR points $\{\mathbb{P}^{(h)}\}_{h=1}^{H}$;\\
\ENSURE ~~\\ 
Level feature for every point $L(\mathbf{P}_i^{(h)})$
\STATE Initial $L(\mathbf{P}_i^{(h)})$ with the zero ($h=1\rightarrow H,i=1\rightarrow N^{(h)} $)
\FOR{$h=1$ to $H$}
\FOR{$i=1$ to $N^{(h)}$}
\IF{$ob(\mathbf{P}_i^{(h)})=1$}
\FOR{$j=i$ to $N^{(h)}$}
\STATE $L(\mathbf{P}_j^{(h)})=L(\mathbf{P}_j^{(h)})+abs(z_{\mathbf{P}_i^{(h)}}-z_{\mathbf{P}_{i-1}^{(h)}})$
\ENDFOR
\ENDIF
\ENDFOR
\ENDFOR
\end{algorithmic}
\end{algorithm}

\subsection{Feature Extraction Based on Initial Drivable Area}
Once $\mathbb{S}_{int}$ is obtained, four features ("level" feature, normal feature, color feature, strength feature) are all calculated superpixel by superpixel to describe $\mathbb{S}_{int}$.
\subsubsection{"Level" Feature}
Our method focuses on detecting the drivable area. Therefore, a feature called "level" is proposed to describe the drivable degree, and Algorithm.\ref{alg:level} shows steps to calculate it. LIDAR points in $\mathbb{P}^{(h)}$ are arranged in accordance with the distances to $\mathbf{P}_{base}$, that is, every point $\mathbf{P}_{i}^{(h)}$ in $\mathbb{P}^{(h)}$ satisfies:
\begin{equation}\label{dist}
dist(\mathbf{P}_{i}^{(h)},\mathbf{P}_{base})>dist(\mathbf{P}_{i-1}^{(h)},\mathbf{P}_{base}),i=2 \rightarrow N^{(h)}
\end{equation}

Then, the "level" feature of superpixel $S_i$ is defined as:
\begin{equation}
L({S}_i)=\frac{1}{||\mathbb{P}_{S_i}||}\displaystyle\sum_{\mathbb{P}_{S_i}} L(\mathbf{P}_i)
\end{equation}
Because $L(\mathbf{P}_i)$ corresponds to $z_{i}$, a small $L(\mathbf{P}_i)$ means a high drivable degree of the relevant area. A probability map is generated in the "level" feature space, where the probability distribution is represented by a Gaussian-like model with parameters ${\mu_{l}}$ and $\sigma_{^{l}}^2$ as:
\begin{equation}\label{eq:level}
L_{prob}({S}_i) = \left\{ \begin{array}{l}
\exp ( - \frac{{{{(L({S}_i) - {\mu_{l}})}^2}}}{{2\sigma_{^{l}}^2}}){\rm{,~if~}}L({S}_i) \ge {\mu_{\l}}\\
1{\rm{~~~~~~~~~~~~~~~~~~~~~~~~~~~otherwise}}
\end{array} \right.
\end{equation}
where $L_{prob}({S}_i)$ is the probability that $S_i$ belongs to the drivable area in the "level" feature space.
The parameter ${\mu_{l}}$ and $\sigma_{^{l}}^2$ can be calculated throughout $\mathbb{S}_{int}$ in a self-learning manner without training.

\subsubsection{Normal Feature}
 The normal feature $N(S_i)$ of each superpixel is designed as the minimum value of the $z_i^n$ among $\mathbb{P}_{S_i}$.
 As mentioned in Section \uppercase\expandafter{\romannumeral3}, $z_i^n$ represents the angle deviation of the relevant spatial triangle from the horizon. Thus, a larger $z_i^n$ means a higher drivable degree of $\mathbf{P}_i$. Namely, the larger $z_i^n$ is, the more flat the area will be. Similar to (\ref{eq:level}), a Gaussian-like model with parameters
${\mu_{n}}$ and $\sigma_{^{n}}^2$ is built as:
\begin{equation}\label{eq:VN}
N_{prob}({S}_i) = \left\{ \begin{array}{l}
\exp ( - \frac{{{{(N({S}_i) - {\mu_{n}})}^2}}}{{2\sigma_{^{n}}^2}}){\rm{,if}}N({S}_i) \le {\mu_{n}}\\
1{\rm{~~~~~~~~~~~~~~~~~~~~~~~~~~~otherwise}}
\end{array} \right.
\end{equation}
where $N_{prob}({S}_i)$ is the probability that $S_i$ belongs to the drivable area in the normal feature space. The estimation of ${\mu_{n}}$ and $\sigma_{^{n}}^2$ is the same as ${\mu_{l}}$ and $\sigma_{^{l}}^2$ mentioned above. Similarly, no manual setting or training is needed.

\subsubsection{Color Feature}
The color feature of $\mathbb{S}_{int}$ is calculated using $I_{ii}$. A Gaussian model with parameters ${\mu_{c}}$ and $\sigma_{^{c}}^2$ is built as:
\begin{equation}
C_{prob}(S_i) = \exp ( - \frac{{{{({I_{ii}(S_i)} - {\mu_{c}})}^2}}}{{2\sigma_{^{c}}^2}})
\end{equation}
where $C_{prob}(S_i)$ is the probability that $S_i$ belongs to the drivable area in the color feature space. And ${I_{ii}(S_i)}$ is the color of $S_i$ in illumination invariant color space. ${\mu_{c}}$ and $\sigma_{^{c}}^2$ are calculated throughout $\mathbb{S}_{int}$ like ${\mu_{l}}$ and $\sigma_{^{l}}^2$.
\subsubsection{Strength Feature}
The number of ray points within superpixel $S_i$ is counted to measure the smoothness of the relevant area, and is defined as strength feature $Sg({{{S}}_i})$. Different from above Gaussian models, the probability that $S_i$ belongs to the drivable area is calculated as:
\begin{equation}
{Sg}_{pro{b_i}}(S_i)={\frac{Sg({{S}_i}){dist(S_i,\mathbf{P}_{base})}}{A(S_i)}}
\end{equation}
where $dist(S_i,\mathbf{P}_{base})$ presents the Euclidean distance between $S_i$ and $\mathbf{P}_{base} $ in image coordinate frame, and $A(S_i)$ presents the area of $S_i$.

\section{Feature Fusion via Belief Prorogation}
\begin{figure}[t]
  \centering
  \includegraphics[width=0.49\textwidth]{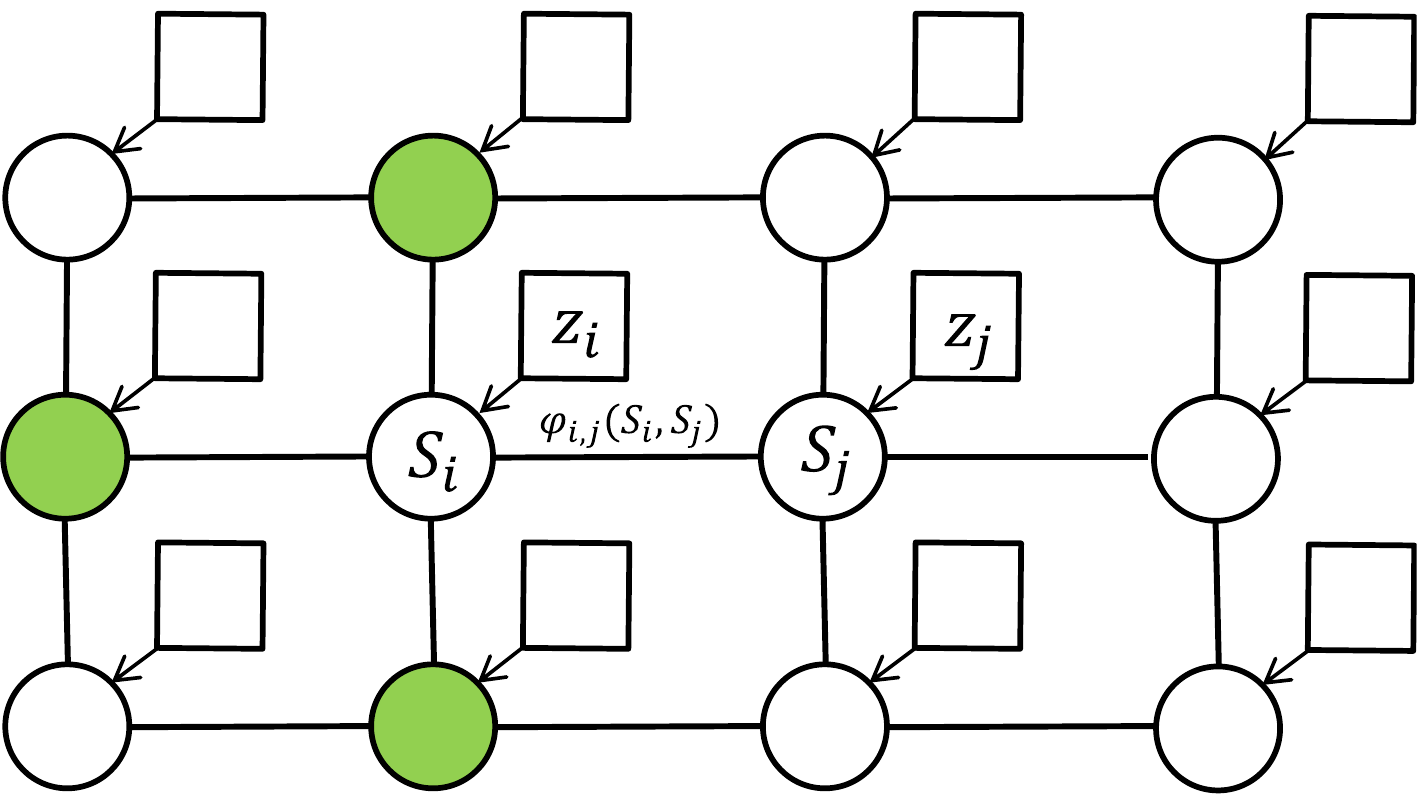}\\
  \caption{The Markov network used to model the positional relationship between adjacent superpixels. A circle node represents the state of a superpixel and a square node represents the observation of the corresponding superpixel.}\label{fig:bp}
\end{figure}
Once all features are obtained as detailed in Section \uppercase\expandafter{\romannumeral4}, these features are fused to get the final results.
The most straightforward fusion method is using Bayesian rule to get the maximum posteriori probability of each superpixel that belongs to the drivable area.
But this fusion method ignores the positional relationship among superpixels which is valuable in this task.
To model this kind of relationship between adjacent superpixels, a Markov network is used.
As shown in \figurename\ref{fig:bp}, a circle node represents the state of a superpixel and a square node represents the observation of the corresponding superpixel. An undirected line models relationship between adjacent superpixels and is calculated by potential compatibility function $\varphi$. A directed line models the observation process.

To perform the inference of the model, the belief propagation algorithm is used\cite{freeman2000learning}.
The local message passing from node $S_i$ to node $S_j$ is:
\begin{equation}\label{eq:message}
m_{i \rightarrow j}=\sum_{S_i}[~p_i(z_i|x_i)\varphi_{i,j}(S_i,S_j)\prod_{k\in \mathcal{N}(S_i) \backslash  j} m_{k \rightarrow i}~]
\end{equation}
where $\mathcal{N}(S_i) \backslash  j$  is the set of of $S_i$'s adjacent superpixels except $S_j$ (green nodes in \figurename\ref{fig:bp}).
The marginal posterior probability of $S_i$ can be obtained by
\begin{equation}\label{eq:bp}
P(S_i|Z) \propto p_i(z_i|S_i)\prod_{j\in \mathcal{N}(S_i) } m_{j \rightarrow i}
\end{equation}
Then the fusion problem is formulated as designing the likehood function $p_i(z_i|S_i)$ and potential compatibility function $\varphi_{i,j}(S_i,S_j)$.
$p_i(z_i|S_i)$ can be obtained by:
\begin{equation}\label{eq:likehood}
\begin{split}
p_i(z_i|S_i)=L_{prob}({S}_i) \ast N_{prob}({S}_i)~~ \\
\ast ~C_{prob}(S_i) \ast {Sg}_{pro{b_i}}(S_i)
\end{split}
\end{equation}
Noticing that $\varphi_{i,j}(S_i,S_j)$ represents the closeness of $S_i$ and $S_j$, so $\varphi_{i,j}(S_i,S_j)$ is defined as:
\begin{equation}\label{eq:phi}
\varphi_{i,j}(S_i,S_j) = \exp ( - \frac{{{{(N({S}_i) - {N({S}_j})})^2}}}{{2\sigma_{^{n}}^2}})
\end{equation}
which is similar to (\ref{eq:VN}).

With (\ref{eq:likehood}) and (\ref{eq:phi}), the $m_{i \rightarrow j}$ is calculated iteratively following (\ref{eq:message}) and the fusion result is then obtained through (\ref{eq:bp}).

\section{Experimental Results And Discussion}
In order to validate our approach, we test it on the ROAD-KITTI benchmark \cite{Fritsch2013ITSC}.
The result is evaluated in BEV with the metrics max F-measure ($\bf MaxF$), average precision ($\bf AP$), precision ($\bf PRE$), recall ($\bf REC$), false positive rate ($\bf FPR$), and false negative rate ($\bf FNR$) for three datasets: Urban Marked (UM), Urban Multiple Marked (UMM), Urban Unmarked (UU).
To show the priority of the proposed algorithm ("Ours Test" in tables below), we compare it with HybridCRF, MixedCRF and LidarHisto, which are the top three methods among LIDAR involved methods from ROAD-KITTI benchmark's websit\footnote{ http://www.cvlibs.net/datasets/kitti/eval\_road.php }.
Besides, our experimental results on training set are also listed below ("Ours Train" in tables below) only to demonstrate that our approach is training-free.

As TABLE \ref{tb:comUM} \ref{tb:comUMM} and \ref{tb:comUU} show, our approach achieves the highest $\bf AP$ in set UM and UU indicating that it is robust for different situations. We also obtain the best score in $\bf PRE$ and $\bf FPR$ in UMM and UU, which indicates that the road areas are covered well by our results. TABLE \ref{tb:comUB} lists the performance across the three datasets and our approach obtains the best $\bf AP$, $\bf PRE$ and $\bf FPR$.
However, our $\bf{FPR}$ value is higher compared with other methods. This can be explained by the fact that the ground truth represents road areas, while our approach detects drivable areas, which contains road as well as other flat area (lawn, transition zones between sidewalk and road).
In practical application, self-driving cars should choose such flat areas as candidate road for emergency, such as avoiding sudden turning vehicles.
Above all, our method is unsupervised and still it gets such competitive performance compared with supervised methods.
\begin{table}[h] 
\centering
\caption{ Comparison on UM (BEV).}
\label{tb:comUM}
\begin{tabular}{c | c | c | c | c | c | c}
\hline
{\bf UM} & {\bf MaxF} & {\bf AP} & {\bf PRE} & {\bf REC} & {\bf FPR} & {\bf FNR}\\
\hline
HybridCRF & \textbf{90.99}  & 85.26  & 90.65   & 91.33   & 4.29   & 8.67  \\
MixedCRF & 90.83   & 83.84   & 89.09   & \textbf{92.64}   & 5.17   & \textbf{7.36}  \\
LidarHisto & 89.87   & 83.03   & \textbf{91.28}   & 88.49   & \textbf{3.85}   & 11.51  \\
Ours Test & 84.96   &\textbf{86.51}   & 79.94   & 90.65   & 10.37   & 9.35 \\
Ours Train &86.34	 &88.17	&82.29	   &90.80	&8.98	&9.20  \\
\hline
\end{tabular}
\end{table}

\begin{table}[h] 
\centering
\caption{ Comparison on UMM (BEV).}
\label{tb:comUMM}
\begin{tabular}{c | c | c | c | c | c | c}
\hline
{\bf UMM} & {\bf MaxF} & {\bf AP} & {\bf PRE} & {\bf REC} & {\bf FPR} & {\bf FNR}\\
\hline
HybridCRF & 91.95   & 86.44   & 94.01   & 89.98   & 6.30   & 10.02  \\
MixedCRF & 92.29   & 90.06   & 93.83   & 90.80   & 6.56   & 9.20  \\
LidarHisto & \textbf{93.32}   & \textbf{93.19}   & \textbf{95.39}   & 91.34   &\textbf{4.85}   & 8.66  \\
Ours Test & 92.22   & 92.23   & 91.70   &\textbf{92.74}   & 9.23   &\textbf{7.26}  \\
Ours Train &92.74	&93.87	 &92.51	&92.96	 &8.20	&7.04  \\
\hline
\end{tabular}
\end{table}

\begin{table}[h]
\centering
\caption{ Comparison on UU (BEV).}
\label{tb:comUU}
\begin{tabular}{c | c | c | c | c | c | c}
\hline
{\bf UU} & {\bf MaxF} & {\bf AP} & {\bf PRE} & {\bf REC} & {\bf FPR} & {\bf FNR}\\
\hline
HybridCRF & \textbf{88.53}   & 80.79   & 86.41   & 90.76   & 4.65   & 9.24  \\
MixedCRF & 82.79   & 69.11   & 79.01   & 86.96   & 7.53   & 13.04  \\
LidarHisto & 86.55   & 81.13   & \textbf{90.71}   & 82.75   & \textbf{2.76}   & 17.25  \\
Ours Test & 83.48   &\textbf{84.75}   & 77.19   &\textbf{90.87}  & 8.75   &\textbf{9.13}  \\
Ours Train & 83.20 	&84.97 	&77.45 	&89.86 	&9.31 	&10.14  \\
\hline
\end{tabular}
\end{table}

\begin{table}[h]
\centering
\caption{ Comparison on URBAN (BEV).}
\label{tb:comUB}
\begin{tabular}{c | c | c | c | c | c | c}
\hline
{\bf URBAN} & {\bf MaxF} & {\bf AP} & {\bf PRE} & {\bf REC} & {\bf FPR} & {\bf FNR}\\
\hline
HybridCRF & \textbf{90.81}   & 86.01   & 91.05   & 90.57   & 4.90   & 9.43    \\
M-CRF & 89.46   & 83.70   & 88.52   & 90.42   & 6.46   & 9.59  \\
LidarHisto & 90.67   & 84.79   & \textbf{93.06}  & 88.41   & \textbf{3.63}   & 11.59  \\
Ours Test & 87.72  & \textbf{87.84}  & 83.97   & \textbf{91.83}   & 9.65  & \textbf{8.17} \\
Ours Train &87.43 	&89.00 	&84.08 	&91.21 	&8.83 	&8.79  \\
\hline
\end{tabular}
\end{table}
Besides, in order to testify how much the feature fusion step boosts the performance, we compare the final results with results from single feature space as well as $\mathbb{S}_{int}$.
All the results in TABLE \ref{tb:selfUB} are obtained from experiments on training set.
As TABLE \ref{tb:selfUB} shows, feature fusion achieves a significant boost in $\bf MaxF$, $\bf AP$ and $\bf PRE$.
It should be noticed that the $\mathbb{S}_{int}$ ("Initial" in the table below) performs outstandingly in $\bf REC$ and ${\bf FNR}$ with a similar ${\bf FPR}$ with Baseline, so that it's reasonable to use $\mathbb{S}_{int}$ to estimate parameters as described in Section \uppercase\expandafter{\romannumeral4}.
\begin{table}[h]
\centering
\caption{ Comparison on URBAN Training Set (BEV).}
\label{tb:selfUB}
\begin{tabular}{c | c | c | c | c | c | c}
\hline
{\bf URBAN} & {\bf MaxF} & {\bf AP} & {\bf PRE} & {\bf REC} & {\bf FPR} & {\bf FNR}\\
\hline
Baseline & 77.95   & 82.47   & 72.83   & 83.88   & 20.03   & 16.11  \\
Initial & 81.23   & 67.64   & 70.75   & \textbf{96.25}   & 21.35   & \textbf{3.75}  \\
Color & 85.35   & 79.76   & 78.93   & 93.43   & 12.81   & 6.57  \\
Strength & 84.16   & 86.37   & 79.36   & 89.70   & 12.76   & 10.30  \\
Level & 86.14  & 76.62   & 80.58   & 92.96   & 11.65   & 7.04  \\
Normal & 87.04   & 79.23   & 83.48   & 91.28   & 8.98   & 8.70  \\
Fusion &\textbf{87.43} 	&\textbf{89.00} 	&\textbf{84.08} 	&91.21 	&\textbf{8.83} 	&8.79  \\
\hline
\end{tabular}
\end{table}

\section{Conclusion And Future Work}
In this paper, an unsupervised approach for detecting drivable areas is proposed that fuses four features via belief prorogation.
Our approach combines both pixel information and depth information to overcome the drawbacks of using single observation when faced with highly various traffic scene and light conditions. Without the need of strong hypothesis, training steps or manually labelled data, our method is proved to be a general approach for self-driving cars.
Besides, the experiments on the ROAD-KITTI benchmark verified the efficiency and robustness of our approach.
In future work, we will first focus on separating road areas from the drivable areas and locating candidate drivable areas for emergency. Besides, a more suitable dataset with hierarchical labels for drivable area is required. Finally, we intend to realize a FPGA implementation of our approach to achieve a real-time application for self-driving cars.


\section*{Acknowledgment}

This research was partially supported by the National Natural Science Foundation of China (No. 61627811, L1522023), the Programme of Introducing Talents of Discipline to University (No. B13043)





\bibliographystyle{IEEEtran}
\bibliography{reflzy}
%
%
%

\end{document}